\PassOptionsToPackage{table}{xcolor}
\documentclass[]{bytedance_seed}

\usepackage[toc,page,header]{appendix}
\usepackage[utf8]{inputenc}
\usepackage[T1]{fontenc}
\usepackage{url}
\usepackage{booktabs}
\usepackage{multirow}
\usepackage{amsmath}
\usepackage{amsfonts}
\usepackage{nicefrac}
\usepackage{microtype}
\usepackage{graphicx}
\usepackage{subcaption}
\usepackage{wrapfig}
\usepackage{tikz}
\usepackage[export]{adjustbox}
\usepackage{placeins}
\usepackage{longtable}
\usepackage{tablefootnote}
\usepackage{caption}

\newcommand{\pizero}{$\pi_0$}
\newcommand{\pizerofast}{$\pi_0$-Fast}
\newcommand{\groot}{GR00T}

\newcommand{\ours}{SG-LP}
\renewcommand{\bold}[1]{\textbf{#1}}

\title{TTT-VLA: Test-Time Latent Prompt Optimization for Vision-Language-Action Models}

\author[1,2,*]{Wenbo Zhang}
\author[1,3]{Jianxiong Li}
\author[4]{Shuai Yang}
\author[1,5]{Sijin Chen}
\author[6]{Jiajun Liu}
\author[2,\dagger]{Lingqiao~Liu}
\author[1,\dagger]{Xiao Ma}

\affiliation[1]{ByteDance Seed}
\affiliation[2]{The University of Adelaide}
\affiliation[3]{Tsinghua University}
\affiliation[4]{Zhejiang University}
\affiliation[5]{The~University of Hong Kong}
\affiliation[6]{CSIRO Data61}

\contribution[*]{Work done at ByteDance Seed}
\contribution[\dagger]{Corresponding authors}

\date{June 1, 2026}
\correspondence{\email{xiao.ma@bytedance.com}, \email{lingqiao.liu@adelaide.edu.au}}

\abstract{
Vision-Language-Action (VLA) models trained on large-scale data have made remarkable progress, but they remain vulnerable to distribution shifts at deployment time. 
    Recent VLA models suggest that prompts can serve as an efficient interface for steering policy behavior, but existing prompt-based steering typically relies on external guidance. This raises a natural question: can test-time training (TTT) for VLA be achieved by optimizing a prompt, so that the steering interface itself can be learned and adapted from interaction? We address this question with TTT-VLA, a test-time training framework based on Latent Prompt Optimization (LPO).
    During training, the latent prompt is learned with an additional proxy task, providing an extra learned conditioning signal for policy learning.
    At test time, TTT is performed by collecting interaction data from the current environment and optimizing only the latent prompt on those data using the proxy task's self-supervised signal, without modifying the policy itself. Experiments on SimplerEnv demonstrate that the proposed method consistently improves task success rates in both single- and multi-embodiment settings. Further analysis shows that the gains arise primarily from correcting a small number of critical decisions rather than globally altering policy behavior. These results suggest that LPO provides an effective and practical pathway for deployment-time improvement of foundation manipulation policies.
}

\begin{document}
\maketitle

\section{Introduction}

Vision-Language-Action (VLA) models are a promising framework for general-purpose robot control because they unify visual perception, language conditioning, and action generation within a single policy.
The pursuit of better generalization in VLA has motivated a range of technical directions, including large-scale pretraining~\cite{black2024pi_0,brohan2022rt,brohan2023rt,kim24openvla,padalkar2023open}, latent action models~\cite{univla_2025,lapa_2025}, and world-model-based methods~\cite{cosmos_policy_2026,dreamzero_2026}. Another practical route toward this goal is to improve the policy at deployment time, especially when the model must act under shifts in visual conditions, environment, or embodiment.
However, the dominant approach to deployment-time improvement is still reinforcement learning (RL), which can improve a policy from interaction but is often costly and reward-dependent~\cite{pi06_2025,grrl_2025}.

At the same time, recent work suggests that prompt-based steering is becoming an important interface for deployment-time improvement in VLA~\cite{zheng2024tracevla,molmoact_2025,zhao2025cotvla,pi07_2026}.
In particular, with human assistance such as factorized instructions, an off-the-shelf policy can achieve better performance. While useful, this paradigm cannot learn from deployment experience. On the other hand, as highlighted by $\pi_{0.7}$, task success and proficiency can also depend on finer-grained context that is difficult to express with text alone, such as episode quality, execution speed, or visual cues~\cite{pi07_2026}. These factors reflect control-relevant temporal and spatial context, suggesting that there is still substantial room to improve how such context is represented for VLA control.

\begin{wrapfigure}{t}{0.48\linewidth}
    \vspace{-20pt}
    \centering
    \includegraphics[width=\linewidth]{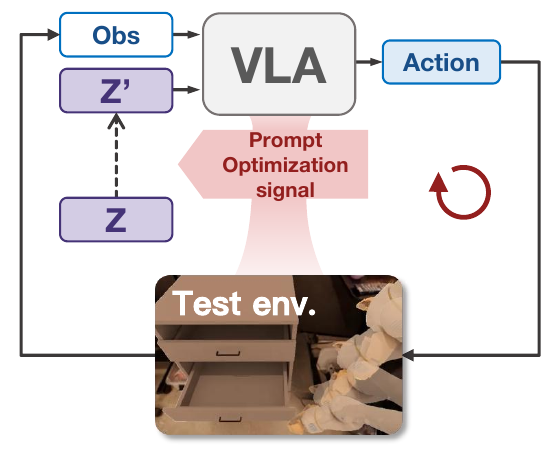}
    \caption{\small{
        Overview of TTT-VLA with Latent Prompt Optimization.
    }}
    \label{fig:teaser}
\end{wrapfigure}

\begin{table*}[b]
    \centering
    \small
    \caption{\textbf{Comparison of deployment-time improvement strategies for VLA.}
    These strategies differ in how they intervene on the policy at deployment time, what gets updated, and what signal drives improvement. TTT-VLA belongs to the self-supervised test-time training regime: it optimizes only a latent prompt while keeping the policy backbone frozen. More comparisons refer to~\ref{app:teaser_comparison}.}
    \label{tab:teaser_comparison}
    \setlength{\tabcolsep}{6pt}
    \renewcommand{\arraystretch}{1.25}
    \resizebox{\textwidth}{!}{
        \begin{tabular}{p{0.13\textwidth}p{0.20\textwidth}p{0.20\textwidth}p{0.24\textwidth}}
            \toprule
            & \textbf{Human-Assisted Steering} & \textbf{RL Post-Training} & \textbf{Test-Time Training } \\
            \midrule
            \textbf{Intervention interface} &
            Prompts, e.g., task-factorized instruction, metadata, trace &
            Typically policy itself &
            Typically parts of the parameters. In this paper, we only use latent prompts. \\
            \textbf{Improvement signal} &
            Human guidance / external signal&
            Reward &
            Self-supervised loss\\
            \textbf{Policy backbone status} &
            Frozen &
            Generally updated &
            Frozen \\
            \bottomrule
        \end{tabular}
    }
    \vspace{-5pt}
\end{table*}

Motivated by this observation, we explore a new direction for deployment-time improvement: learning a latent prompt $z$ during training to condition a policy, and then improving the policy at test time by adapting only $z$.
We address this problem with TTT-VLA, a framework based on \emph{Latent Prompt Optimization}. As shown in Fig.~\ref{fig:teaser}, the core idea is to obtain an improvement signal from interaction with the test environment and use it to optimize the latent prompt. Because the VLA is trained to condition on this prompt, test-time knowledge can be absorbed through a lightweight interface without modifying the policy backbone.
In our current implementation, the improvement signal comes from a proxy task called \emph{state grounding}, which uses observation-state supervision to shape the latent prompt. This proxy task encourages the latent prompt to capture spatially relevant information~\cite{rosa_2025}, and we further design training strategies to strengthen this effect. The resulting latent prompt serves two roles: during training, it provides an additional learned conditioning signal for policy learning; at test time, interaction data from the current environment are collected and used to update only the latent prompt of a frozen policy.

We evaluate this idea in SimplerEnv~\cite{simpler_2025} under both single-embodiment and multi-embodiment settings. Our evaluation follows the official protocol, in which the policy is trained on real robot data and tested in simulation, naturally introducing a deployment-time domain gap.
In the single-embodiment setting, we evaluate on WidowX and Google Robot tasks.
In the multi-embodiment setting, we train on BridgeData V2~\cite{walke2023bridgedata} of OXE-Aug~\cite{oxe_auge_2025} with nine embodiments and then evaluate on WidowX.
Across these settings, we achieve consistent improvements with TTT-VLA while keeping the main policy frozen.
The analysis further suggests that much of the gain comes from what we call \emph{critical decision steering}, where test-time training corrects a small number of key actions rather than globally rewriting the full trajectory.
Our contributions are summarized as follows:
\begin{itemize}
    \item TTT-VLA, a deployment-time improvement framework for VLA based on Latent Prompt Optimization, which learns a latent prompt during training and performs prompt-only test-time training on a frozen policy at deployment time.
    \item Consistent deployment-time improvements in both single-embodiment and multi-embodiment settings, showing that a latent prompt can serve as an effective interface for learning from interaction.
    \item An investigation of effective training and update strategies for the latent prompt, revealing that the gains mainly arise from a phenomenon we term critical decision steering.
\end{itemize}

\section{Related Work}

\subsection{Deployment-Time Improvement for Manipulation Policies}
Deployment-time improvement for manipulation policies has been studied through RL post-training, human-assisted prompt steering, and, more recently, test-time training.
One established line of work improves VLA policies through reinforcement learning from deployment experience.
Representative examples include $\pi^{*}_{0.6}$, which studies RL-based self-improvement from heterogeneous experience and human corrections, and GR-RL, which turns a generalist VLA into a dexterous specialist through multi-stage reinforcement learning \citep{pi06_2025,grrl_2025}.

Another line of work uses the VLA conditioning interface itself as a channel for human assistance at deployment time.
Yell At Your Robot uses real-time language corrections during execution, while MolmoAct and TraceVLA expose editable traces as structured steering signals for spatial-temporal guidance \citep{yellrobot_2024,molmoact_2025,zheng2024tracevla}.
\(\pi_{0.7}\) broadens this interface further with metadata, language guidance, and image-goal context, showing that human- or externally-specified context can improve trajectory quality and generalization to unseen tasks \citep{pi07_2026}.

Early work on test-time training for manipulation policies includes PAD \citep{pad_2020}, which showed on small-scale policies that self-supervised signals can help manipulation policies adapt to environmental changes. However, PAD was not studied in the foundation-model regime. As a result, neither its design nor its conclusions transfer directly to modern VLA models. More recently, test-time training has begun to emerge in the VLA literature. A contemporaneous work is WorldAgen \citep{worldagen_2026}, which jointly learns action prediction and world modeling, and then adapts its world-model branch from newly collected transitions at deployment time. However, its test-time training scheme is built around a tailored architecture that is introduced together with the method, rather than around a strong state-of-the-art VLA model. By contrast, we study prompts as a lightweight adaptation interface for pretrained VLA models, so that test-time training is performed on top of a strong foundation policy with already competitive performance.

\subsection{Prompt-Conditioned VLA}
Recent VLA models are increasingly organized around prompt-conditioned interfaces that extend beyond language alone.
Early systems such as RT-2, OpenVLA, and Octo established language-conditioned robot control, and Octo further showed that subgoal images can serve as an additional task-specification channel \citep{brohan2023rt,kimOpenVLAOpenSourceVisionLanguageAction2024,octo_2023,yang2025instructvla}.
Recent work further broadens prompt by using structured trajectory or trace inputs as prompts, as in MolmoAct and TraceVLA \citep{molmoact_2025,zheng2024tracevla}.
The \(\pi\) series further expands this line: \(\pi_0\) improves the breadth and dexterity of language-conditioned control, \(\pi_{0.5}\) augments the prompt with hierarchical sub-task instructions, and \(\pi_{0.7}\) further broadens the conditioning space with metadata and image-goal context derived from a world model \citep{black2024pi_0,pi05_2025,pi07_2026}.
Together, these works show that richer prompt interfaces are central to modern VLA performance.
Our method is complementary to this line: rather than adding another human-specified prompt channel, we learn a latent prompt that captures control-relevant context for a VLA policy and can be updated with test-time training.

\section{Method: Latent Prompt Optimization}
In general, a VLA policy can be written as a conditional policy that takes the current observation $o$ together with a conditioning context $c$ and predicts an action chunk $a$ through
\begin{equation}
    a \sim \pi_\theta(a \mid o, c),
    \label{eq:policy_overview}
\end{equation}
where $c$ is a unified conditioning variable.
It may include explicit conditions with readable semantics, such as language, state, image, or other structured context, as well as an implicit condition in the form of a learnable latent prompt $z$. In our current implementation, $z \in \mathbb{R}^{n \times d}$ is instantiated as a learnable block of $n$ prompt tokens, where $d$ is the hidden dimension of the policy backbone and $n$ is a hyperparameter controlling the prompt length.
Following flow-based VLA formulations, we model action generation as conditional flow matching over an action chunk.
Let $a_0 \sim p_0$ denote a noise sample in action space and let $a_1 = a$ denote the target action chunk.
For an interpolation time $t \sim \mathcal{U}(0,1)$, we define the noised action trajectory as $a_t = (1 - t) a_1 + t a_0$ and the corresponding target vector field as $u_t = a_0 - a_1$.
The policy predicts the vector field conditioned on the current inputs and the full conditioning context:
\begin{equation}
    \mathcal{L}_{\mathrm{act}}(\theta, z)
    =
    \mathbb{E}_{(o,c,a),\,a_0,\,t}
    \left[
        \left\|
            v_\theta(a_t, t; o, c) - u_t
        \right\|_2^2
    \right].
    \label{eq:action_loss}
\end{equation}
This objective defines the base policy learning problem in our framework.
Latent Prompt Optimization does not replace the action objective.
Instead, it adds a learned conditioning variable that can be optimized during training and later adapted at deployment time.
The following subsection introduces how this latent prompt is updated at deployment time through an abstract proxy objective, and the later implementation subsection specifies the concrete proxy used in our current system.

\begin{figure}[t]
    \centering
    \includegraphics[width=\linewidth]{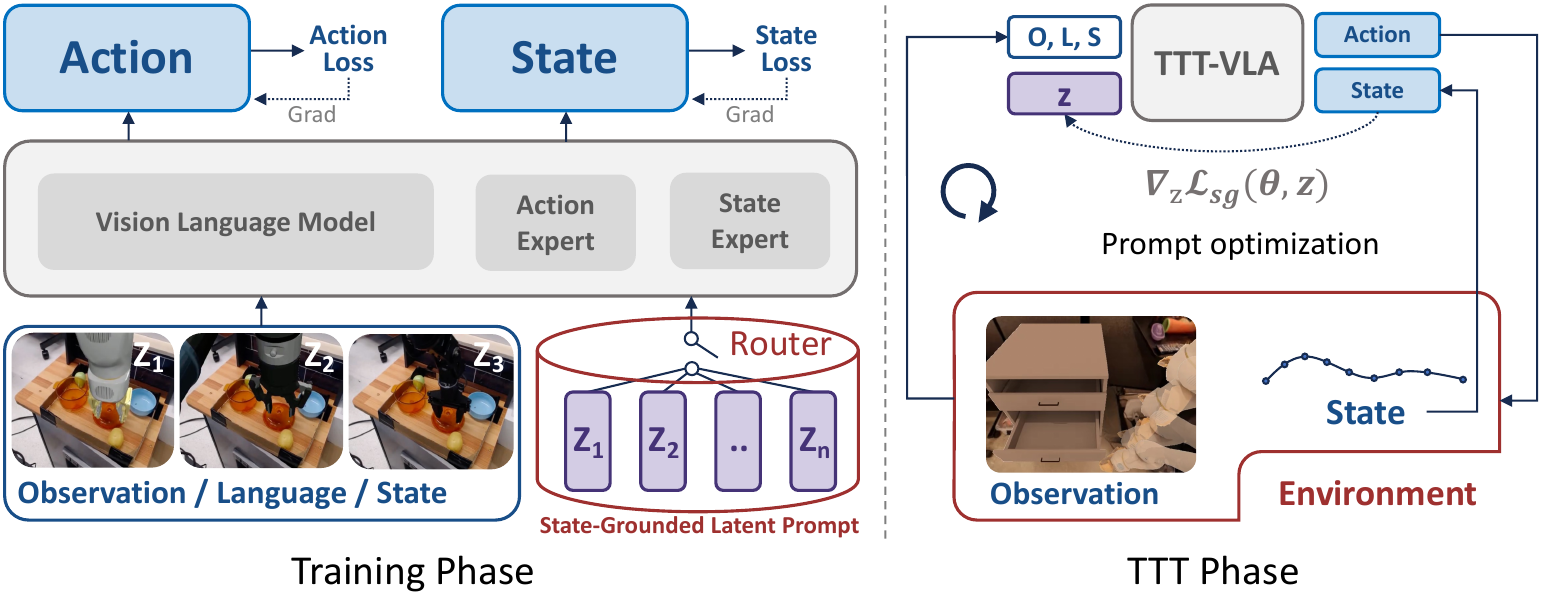}
    \caption{Overview of TTT-VLA.
    The policy is conditioned on the current observation, the explicit prompt, and a latent prompt.
    During training, the latent prompt is optimized jointly with the policy and with the state grounding proxy task.
    During deployment, test-time training updates only the latent prompt from interaction with the environment without modifying the policy itself.}
    \label{fig:method_overview}
    \vspace{-10pt}
\end{figure}

\subsection{Test-Time Training via Latent Prompt Optimization}

Inspired by prior test-time training and deployment-time self-supervised adaptation works~\cite{ttt_2020,ttt_mae_2022,pad_2020}, we introduce a proxy task to enable test-time training for latent prompt optimization.
Since the action objective is not directly available for prompt-only adaptation at deployment time, this proxy task provides the auxiliary learning objective needed to carry out the test-time training procedure.
At the same time, it couples the latent prompt to control-relevant context during training so that the same objective can later be reused for deployment-time adaptation.
During training, we optimize the policy with both the action objective and the proxy objective so that the latent prompt is useful for action prediction while remaining adaptable through the same proxy signal:
\begin{equation}
    \mathcal{L}_{\mathrm{train}}(\theta, z)
    =
    \mathcal{L}_{\mathrm{act}}(\theta, z)
    +
    \mathcal{L}_{\mathrm{proxy}}(\theta, z).
    \label{eq:joint_train_loss}
\end{equation}

At test time, our primary procedure is prompt-only adaptation from a current-environment interaction buffer.
We first collect interaction data from the current environment, organize the resulting samples into a buffer, and then update only the latent prompt while keeping the policy backbone frozen.
The resulting test-time training rule can be written as
\begin{equation}
    z \leftarrow z - \eta \nabla_z \mathcal{L}_{\mathrm{proxy}},
    \label{eq:ttt_update}
\end{equation}
using interaction data from the current environment.
Here, $\mathcal{L}_{\mathrm{proxy}}$ denotes the same proxy objective introduced in training, and its concrete instantiation is given later as the state grounding task in Sec.~\ref{sec:state_grounding_proxy}.
This preserves a clean division of labor: the action expert remains responsible for action generation, while the latent prompt is the only variable adapted during deployment.
We also explore online prompt updates as an ablation, but in our current experiments this mode is unstable and the prompt representation tends to collapse, so the main method uses the offline buffer formulation above.

\subsection{Architecture and Training Strategy}

\paragraph{Network Structure.} As illustrated in Fig.~\ref{fig:method_overview}, our current implementation instantiates TTT-VLA with a Mixture-of-Transformers (MoT) architecture~\cite{black2024pi_0} composed of two flow-matching experts: an action expert that predicts the flow field in action space, and a state-grounding expert that predicts the flow target in state space for robot state grounding. We consider a hard router for latent prompt selection. If the training data contains multiple embodiments, multiple latent prompts are randomly initialized. In the single-embodiment setting, this design naturally degenerates to a single latent prompt.
\begin{wrapfigure}{r}{0.48\linewidth}
    \centering
    \includegraphics[width=\linewidth]{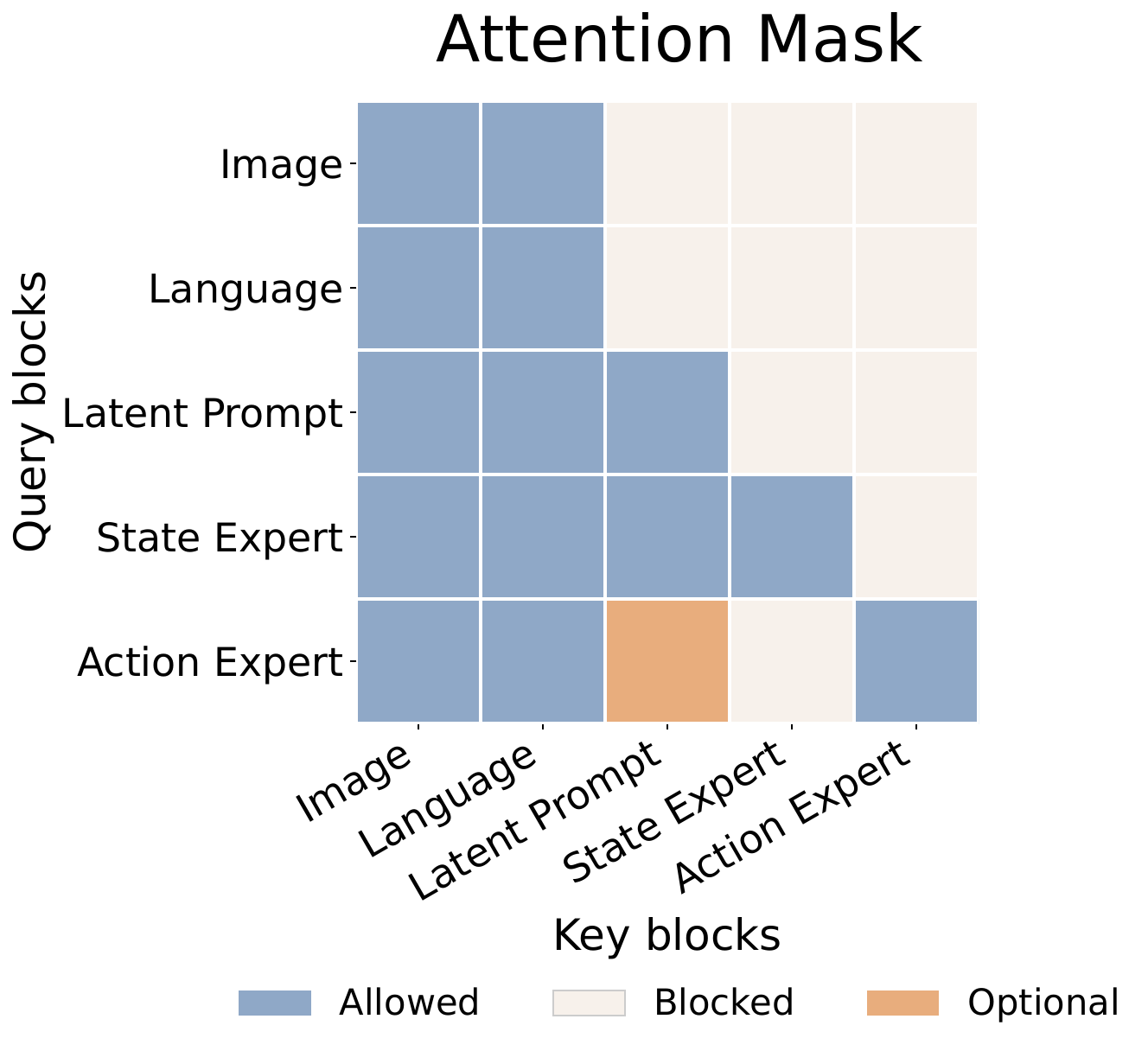}
    \caption{\small{
        Attention mask used in the MoT network.
    }}
    \label{fig:attention_mask}
    \vspace{-10pt}
\end{wrapfigure}

The detailed attention mask is shown in Fig.~\ref{fig:attention_mask}.
Whether the latent prompt can be attended to by the action expert depends on specific training strategies, which are described later. During training, parameters are initialized from the pretrained $\pi_{0.5}$ checkpoint. Note that there is a slight difference in the state input: it is moved from language-form conditioning into the action expert, following the implementation of $\pi_{0}$. This change is mainly introduced to avoid conflict with our proxy task. Our main comparisons focus on two variants.
\paragraph{State Grounding as the Proxy Task.}
\label{sec:state_grounding_proxy} 
In our current implementation, Latent Prompt Optimization is instantiated with a proxy task called \emph{state grounding}, in which a dedicated expert predicts the robot end-effector position and gripper state from the current observation, the explicit prompt, and the latent prompt.
Let $s_1 = s$ denote the target robot state, which consists of the end-effector position and gripper state, and let $s_0 \sim p_0^{s}$ denote a noise sample in state space.
For an interpolation time $t \sim \mathcal{U}(0,1)$, we define the noised state trajectory as $s_t = (1 - t) s_1 + t s_0$ and the corresponding target vector field as $u_t^{s} = s_0 - s_1$.
The state-grounding expert predicts this vector field conditioned on the current inputs and the full conditioning context:
\begin{equation}
    \mathcal{L}_{\mathrm{proxy}}(\theta, z)
    =
    \mathcal{L}_{\mathrm{sg}}(\theta, z)
    =
    \mathbb{E}_{(o,c,s),\,s_0,\,t}
    \left[
        \left\|
            v_{\theta}^{s}(s_t, t; o, c) - u_t^{s}
        \right\|_2^2
    \right].
\end{equation}
We use this task as the proxy objective for two reasons.
First, these state targets are available at both training and deployment time without extra annotation cost, making them a practical self-supervised signal.
Second, accurate state prediction requires spatial understanding of the scene together with embodiment-specific understanding of the robot, and therefore provides rich control-relevant
cues~\cite{rosa_2025,worldagen_2026}.
The remaining question is how to ensure that these cues are more effectively absorbed into the latent prompt rather than other parameters, which motivates the training strategy below.
\paragraph{Training Strategy for Single- and Multi-Embodiment Settings.}
In the multi-embodiment setting, embodiment provides a natural partition of heterogeneous robot data into distinct control-relevant contexts.
We therefore associate different state-grounded latent prompts with different embodiments and jointly optimize them with both the action objective and the state grounding objective.
This design encourages cross-embodiment control differences to be absorbed into the latent prompt, rather than being entirely entangled in shared model parameters.

The single-embodiment setting is more data-limited, so this natural partition becomes much weaker and prompt learning must be enforced more explicitly.
To address this issue, we use a two-stage training strategy.
As indicated by Fig.~\ref{fig:attention_mask}, the connection from the latent prompt to the action expert is optional: in the first stage, it is blocked with probability $1$, and in
the second stage, it is blocked with probability $0.5$.
The ablation of this random-drop design is provided in Appendix Figure~\ref{fig:random_drop_ablation}. Without random drop, simply adding a latent prompt can already degrade the base policy performance, suggesting that stronger decoupling is important in the single-embodiment regime. Related prompt-drop or conditioning-drop strategies have also been used in prior work~\cite{pi07}.
In addition, we constrain the gradient flow of the state-grounding objective to the state expert and the latent prompt, while the gradient flow of the action objective is restricted to
the VLM backbone and the action expert.
Together, these constraints encourage the latent prompt to absorb control-relevant context from the proxy task before it is more strongly coupled to action prediction.

\begin{table*}[!h]
    \centering
    \small
    \caption{Performance comparison with state-of-the-art policies on the \textbf{SimplerEnv WidowX} benchmark. For the last three rows, all are initialized from the pretrained $\pi_{0.5}$ checkpoint. Here, `$\pi_{0.5}$' denotes our internal baseline, `$\pi_{0.5}$ + \ours' adds state grounding with latent prompt training, and `$\pi_{0.5}$ + \ours~+
    TTT' further applies test-time training to the learned latent prompt.}
    \label{tab:benchmark_simpler}
    \vspace{0.25ex}
    \setlength{\tabcolsep}{12pt}
        \begin{tabular}{lccccc}
            \toprule
            \bf WidowX Benchmark & \bf Carrot & \bf Eggplant & \bf Spoon & \bf Cube & \bf Overall \\
            \midrule
            RT-1-X~\citep{brohanRT1RoboticsTransformer2023} & 4.2\% & 0\% & 0\% & 0\% & 1.1\% \\
            OpenVLA~\citep{kim24openvla} & 0\% & 4.1\% & 0\% & 0\% & 1.0\% \\
            SpatialVLA~\citep{qu2025spatialvla} & 20.8\% & 70.8\% & 20.8\% & 25.0\% & 34.4\% \\
            Magma~\citep{yang2025magma} & 29.2\% & 91.7\% & 37.5\% & 20.8\% & 44.8\% \\
            Octo-Base~\citep{octo_2023} & 8.3\% & 43.1\% & 12.5\% & 0\% & 16.0\% \\
            Octo-Small~\citep{octo_2023} & 9.7\% & 56.9\% & 47.2\% & 4.2\% & 29.5\% \\
            RoboVLM~\citep{li2024robovlm} & 20.8\% & 79.2\% & 45.8\% & 4.2\% & 37.5\% \\
            InstructVLA~\citep{yang2025instructvla} & 40.3\% & \textbf{94.4\%} & 43.1\% & 9.7\% & 46.9\% \\

            $\pi_{0}$~\citep{black2024pi_0} & 36.1\% & 81.9\% & 45.8\% & 26.4\% & 47.6\% \\
            CogACT~\citep{li2024cogact} & 37.5\% & 91.7\% & 58.3\% & 20.8\% & 52.1\% \\
            ThinkAct~\citep{huang2025thinkact} & 37.5\% & 70.8\% & 58.3\% & 8.7\% & 43.8\% \\
            \midrule
            $\pi_{0.5}$ & \textbf{79.2}\% & 75.0\% & 33.3\% & 16.7\% & 51.1\% \\
            $\pi_{0.5}$ + \ours & 69.5\% & 70.5\% & \textbf{72.5\%} & 41.5\% & 63.5\% \\
            $\pi_{0.5}$ + \ours~+ TTT & 74.5\% & 76.0\% & 71.0\% & \textbf{48.0\%} & \textbf{67.4\%} \\
            \bottomrule
        \end{tabular}
    \setlength{\tabcolsep}{6pt}
    \vspace{-10pt}
\end{table*}

\begin{table*}[t]
    \centering
    \small
    \caption{\textbf{Performance comparison on SimplerEnv Google Robot tasks.}
    Results are reported for the visual-matching and variant-aggregation protocols.}
    \label{tab:simplerenvgooglerobot}
    \setlength{\tabcolsep}{10pt}
    \begin{tabular}{lcccc}
      \toprule
      \multicolumn{5}{c}{\textbf{Visual Matching}} \\
      \midrule
      Model & Pick Coke Can & Move Near & Open/Close Drawer & Avg \\
      \midrule
      HPT~\citep{wang2024hpt} & 56.0\% & 60.0\% & 24.0\% & 46.0\% \\
      TraceVLA \citep{zheng2024tracevla} & 28.0\% & 53.7\% & 57.0\% & 42.0\% \\
      RT-1-X \citep{brohan2022rt} & 56.7\% & 31.7\% & 59.7\% & 53.4\% \\
      Octo-Base \citep{octo_2023} & 17.0\% & 4.2\% & 22.7\% & 16.8\% \\
      OpenVLA \citep{kim24openvla} & 16.3\% & 46.2\% & 35.6\% & 27.7\% \\
      RoboVLM~\citep{li2024robovlm} (zero-shot) & 72.7\% & 66.3\% & 26.8\% & 56.3\% \\
      RoboVLM~\citep{li2024robovlm} (fine-tuned) & 77.3\% & 61.7\% & 43.5\% & 63.4\% \\
      Emma-X~\citep{li2024emmax} & 2.3\% & 3.3\% & 18.3\% & 8.0\% \\
      \pizero~\citep{black2024pi_0} (fine-tuned) & 72.7\% & 65.3\% & 38.3\% & 58.7\% \\
      \pizerofast~\citep{pertsch2025fast} (fine-tuned) & 75.3\% & 67.5\% & 42.9\% & 61.9\% \\
      \groot~\citep{gr00t_n1_2025} (fine-tuned) & 69.3\% & 68.7\% & 35.8\% & 52.4\% \\
      \midrule
      $\pi_{0.5}$ & 84.0\% & 59.2\% & 59.3\% & 67.5\% \\
      $\pi_{0.5}$ + \ours & \textbf{85.0\%} & 66.2\% & 55.6\% & 68.9\% \\
      $\pi_{0.5}$ + \ours~+ TTT & \textbf{85.0\%} & \textbf{71.7\%} & \textbf{60.6\%} & \textbf{72.4\%} \\
      \bottomrule
    \end{tabular}
  
    \vspace{0.75ex}
  
    \begin{tabular}{lcccc}
      \toprule
      \multicolumn{5}{c}{\textbf{Variant Aggregation}} \\
      \midrule
      Model & Pick Coke Can & Move Near & Open/Close Drawer & Avg \\
      \midrule
      TraceVLA \citep{zheng2024tracevla} & 60.0\% & 56.4\% & 31.0\% & 45.0\% \\
      RT-1-X \citep{brohan2022rt} & 49.0\% & 32.3\% & 29.4\% & 39.6\% \\
      Octo-Base \citep{octo_2023} & 0.6\% & 3.1\% & 1.1\% & 1.1\% \\
      OpenVLA \citep{kim24openvla} & 54.5\% & 47.7\% & 17.7\% & 39.8\% \\
      RoboVLM~\citep{li2024robovlm} (zero-shot) & 68.3\% & 56.0\% & 8.5\% & 46.3\% \\
      RoboVLM~\citep{li2024robovlm} (fine-tuned) & 75.6\% & 60.0\% & 10.6\% & 51.3\% \\
      Emma-X~\citep{li2024emmax} & 5.3\% & 7.3\% & 20.5\% & 11.0\% \\
      \pizero~\citep{black2024pi_0} (fine-tuned) & 75.2\% & 63.7\% & 25.6\% & 54.8\% \\
      \pizerofast~\citep{pertsch2025fast} (fine-tuned) & 77.6\% & \textbf{68.2\%} & 31.3\% & 59.0\% \\
      \groot~\citep{gr00t_n1_2025} (fine-tuned) & 46.7\% & 62.9\% & 17.5\% & 43.7\% \\
      \midrule
      $\pi_{0.5}$ & \textbf{82.0\%} & 47.7\% & 44.7\% & 58.1\% \\
      $\pi_{0.5}$ + \ours & 81.7\% & 51.2\% & 42.9\% & 58.6\% \\
      $\pi_{0.5}$ + \ours~+ TTT & 79.3\% & 55.2\% & \textbf{45.8\%} & \textbf{60.1\%} \\
      \bottomrule
    \end{tabular}
    \setlength{\tabcolsep}{6pt}
    \vspace{-5pt}
  \end{table*}

\section{Experiments}

\bold{Benchmark.}
We use SimplerEnv\cite{simpler_2025} as the evaluation benchmark throughout this section. Within this benchmark, we study both single-embodiment and multi-embodiment settings: the former covers fixed-embodiment WidowX and Google Robot tasks, while the latter studies training on OXE-Aug, an augmented Open X-Embodiment (OXE) training set, and evaluation on WidowX. For readability, the WidowX tasks are abbreviated as Spoon, Carrot, Cube, and Eggplant in Tables~\ref{tab:benchmark_simpler}, \ref{tab:simplerenv_bridge_multiembodiment}, \ref{tab:paired_episode_shift}, and Figure~\ref{fig:ttt_analysis}, corresponding to \emph{Put Spoon on Towel}, \emph{Put Carrot on Plate}, \emph{Stack Blocks}, and \emph{Put Eggplant in Basket}, respectively.

\bold{Baseline.}\label{para:baseline}
For comparison, we report public state-of-the-art results from a range of prior methods on SimplerEnv.
As a strict control for our method, we include a baseline without the state expert or the latent prompt, which is a version directly fine-tuned from the original $\pi_{0.5}$.
The second notation is `$\pi_{0.5}$ + SG-LP' (state grounded latent prompt), which extends the baseline with SG-LP training.
The third is `$\pi_{0.5}$ + SG-LP + test-time training (TTT)', which further applies test-time training to the learned latent prompt.
This latter variant is our full method, TTT-VLA.

\bold{Training.}
All models are initialized from the pretrained $\pi_{0.5}$ checkpoint. We move the state input from language-form conditioning into the action expert, following the implementation of $\pi_{0}$, to avoid conflict with the proxy task. In the single-embodiment setting, models are trained for 20,000 steps; in the multi-embodiment setting, they are trained for 40,000 steps. We use AdamW with a learning rate of $1\times10^{-4}$ and a batch size of 1024. All training runs are conducted on 32 NVIDIA H100 GPUs, with a wall-clock time of approximately 20 hours per run.

\bold{Test-Time Training.}
At test time, we update only the latent prompt, initializing it from the pretrained prompt learned during training. We use a learning rate of $1\times10^{-5}$ and a batch size of 128. Unless otherwise stated, we run 500 optimization steps for WidowX and 1000 optimization steps for Google Robot tasks. We do not select the best checkpoint based on evaluation performance; instead, we fix the optimization horizon in advance for each benchmark so that the procedure remains uniform within each setting and stops well after the proxy loss has largely plateaued. Test-time training is conducted on 8 NVIDIA H100 GPUs and typically requires 15--30 minutes of wall-clock time. More details are provided in Appendix~\ref{tab:ttt_config}.

\begin{wrapfigure}{tr}{0.48\linewidth}
    \centering
    \includegraphics[width=0.48\textwidth]{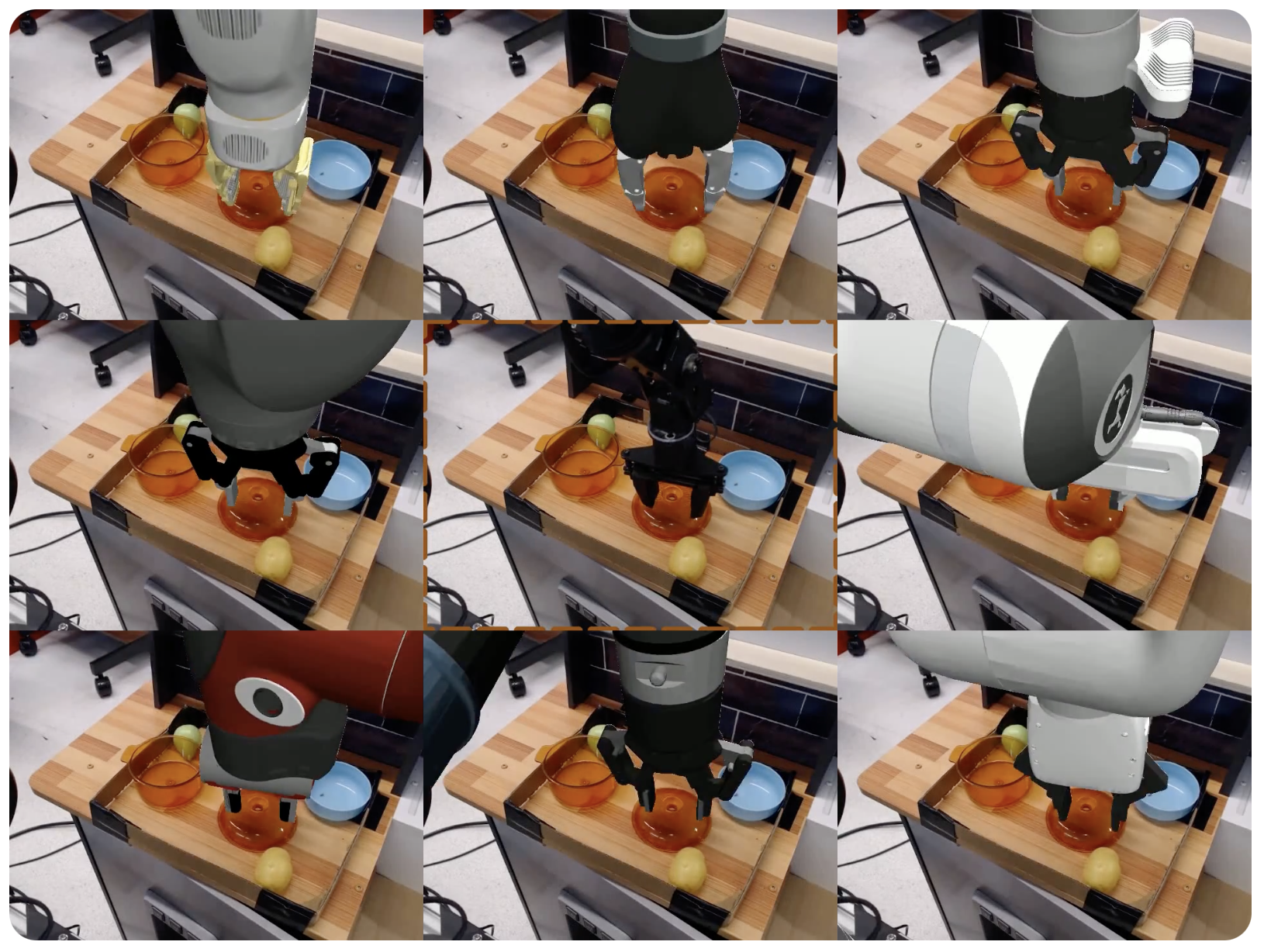}
    \caption{\small{
        OXE-Aug Bridge V2 examples across the nine training embodiments.
    }}
    \label{fig:nine_embodiments}
    \vspace{-10pt}
\end{wrapfigure}

\bold{Evaluation protocol.}
Evaluation in SimplerEnv is stochastic because (1) actions are generated through flow-matching sampling, and (2) the WidowX benchmark has only 24 configurations across four tasks with different object positions and orientations. A single sweep is therefore too noisy for reliable comparison.
Following GR00T~\cite{gr00t_n1_2025}, we therefore evaluate each WidowX task for 200 episodes by cycling through the 24 configurations while allowing sampling noise to produce different actions across trials. Across model variants, we fix only the initial seed rather than the full downstream random stream, so the sampled noise and resulting actions still differ. This reduces evaluation variance and supports more reliable comparison. For Google Robot visual matching and variant aggregation, we use the official default evaluation counts.
After optimizing the evaluation pipeline, the wall-clock time of one full evaluation is approximately 40 minutes for Bridge, 50 minutes for Fractal visual matching, and 3 hours for Fractal variant aggregation.
  
\begin{table*}[t]
    \centering
    \small
    \caption{\textbf{Multi-embodiment evaluation on SimplerEnv WidowX tasks.} The policy is trained on OXE-Aug Bridge V2 with nine embodiments in total and evaluated on four Bridge V2 WidowX tasks.}
    \label{tab:simplerenv_bridge_multiembodiment}
    \setlength{\tabcolsep}{12pt}
        \begin{tabular}{lccccc}
            \toprule
            Method & Carrot & Eggplant & Spoon & Cube & Mean \\
            \midrule
            $\pi_{0.5}$ & 28.0\% & 17.5\% & 43.0\% & \textbf{2.5\%} & 22.8\% \\
            $\pi_{0.5}$ + \ours & 41.0\% & 29.5\% & 43.0\% & 0.5\% & 28.5\% \\
            $\pi_{0.5}$ + \ours~+ TTT & \textbf{43.0\%} & \textbf{34.0\%} & \textbf{48.5\%} & 1.0\% & \textbf{31.6\%} \\
            \bottomrule
        \end{tabular}
    \setlength{\tabcolsep}{6pt}
    \vspace{-5pt}
\end{table*}

\subsection{Single-Embodiment Setting}
We first study the single-embodiment setting, where the main question is whether a learned latent prompt is useful even when the embodiment identity itself is fixed.
Table~\ref{tab:benchmark_simpler} shows the results on the SimplerEnv WidowX benchmark.
Adding Latent Prompt Optimization to $\pi_{0.5}$ improves the mean success rate from 51.1\% to 63.5\%, and test-time training further improves it to 67.4\%.
We observe the same trend on Google Robot tasks in Table~\ref{tab:simplerenvgooglerobot}.
On the visual-matching split, the mean success rate improves from 67.5\% for $\pi_{0.5}$ to 68.9\% with Latent Prompt Optimization, and then to 72.4\% after test-time training.
On the variant-aggregation split, the same sequence improves from 58.1\% to 58.6\% and then to 60.1\%.

\subsection{Multi-Embodiment Setting}
We next study the multi-embodiment setting.
Here, the training data come from OXE-Aug Bridge V2, which contains nine embodiments in total: the original WidowX together with eight additional embodiments, namely Panda, UR5e, Xarm7, Google Robot, Sawyer, Kinova3, IIWA, and Jaco, as shown in Fig.~\ref{fig:nine_embodiments}.
During evaluation, the router selects the corresponding prompt for WidowX before inference.
Results are shown in Table~\ref{tab:simplerenv_bridge_multiembodiment}.
Compared with the $\pi_{0.5}$ baseline, Latent Prompt Optimization improves the mean success rate from 22.8\% to 28.5\%, and test-time training further improves it to 31.6\%.
The improvements are concentrated on Carrot and Eggplant, while Cube remains difficult for all variants. Empirically, we find that Cube is especially hard because it requires more precise grasping and alignment during placement than the other tasks, and this challenge becomes more severe when training on heterogeneous multi-embodiment data.

\begin{figure}[t]
    \centering
    \includegraphics[width=0.95\linewidth]{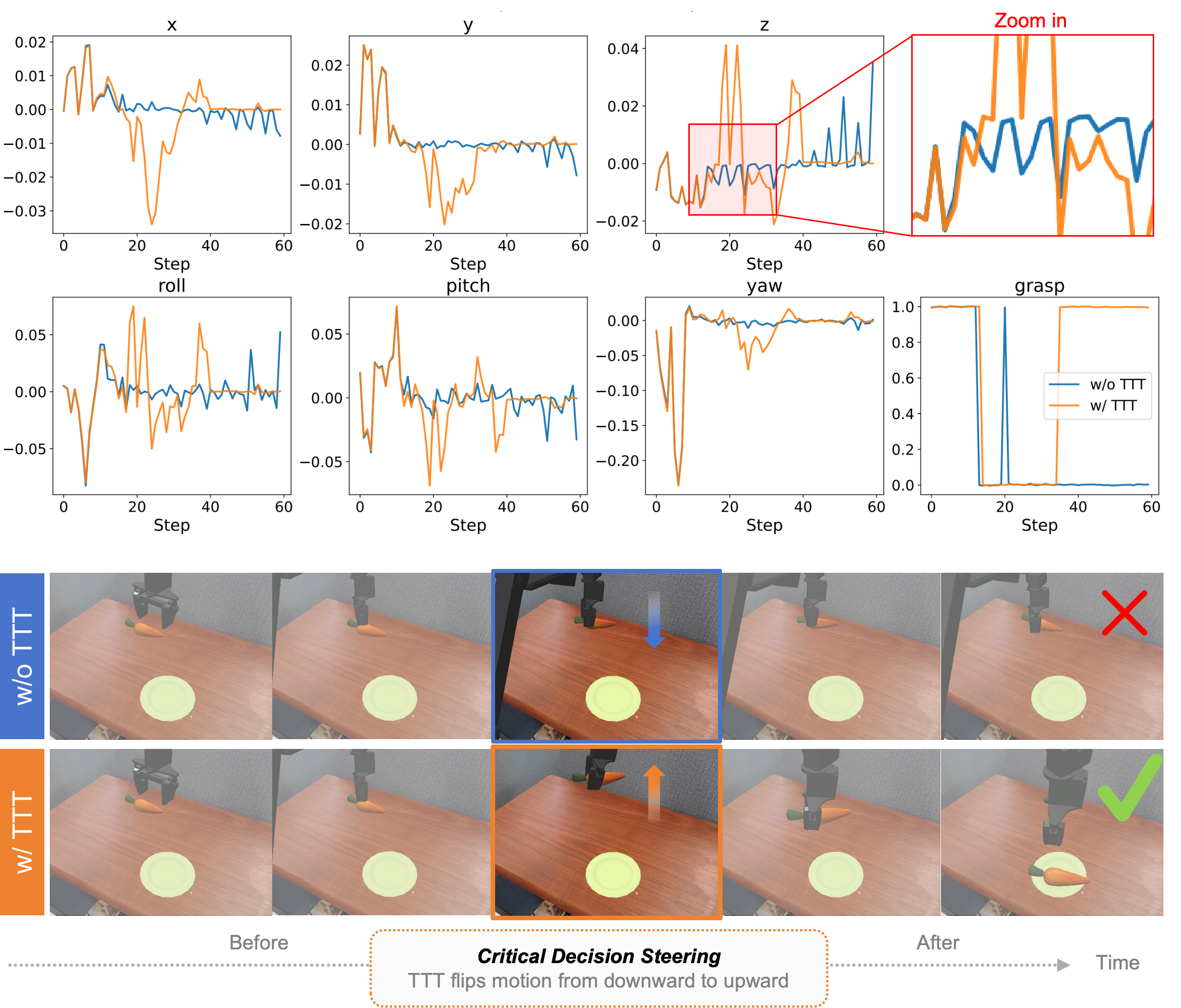}
    \caption{Qualitative visualization of the test-time training effect. The top row compares action trajectories with and without TTT, and the bottom row shows a representative manipulation sequence in which TTT steers a critical decision from a downward motion to an upward corrective motion, leading to task success.}
    \label{fig:ttt_critical_decision}
    \vspace{-10pt}
\end{figure}

\subsection{Ablation and Analysis}
\begin{figure*}[t]
    \centering
    \includegraphics[width=\textwidth]{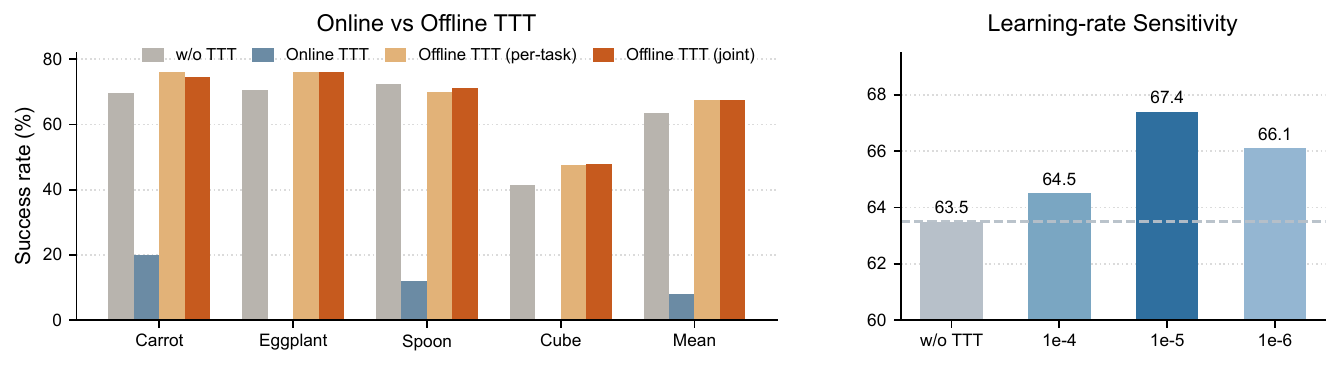}
    \caption{\textbf{Analysis of test-time training configurations in the single-embodiment setting.} Left: task-wise comparison of online and offline TTT strategies, including offline per-task and offline joint variants. Right: learning-rate sensitivity for offline joint TTT, where a moderate learning rate of $10^{-5}$ gives the best mean success rate.}
    \label{fig:ttt_analysis}
\end{figure*}

\bold{Online vs Offline TTT}
Here, Online TTT refers to continuously updating the latent prompt as interaction samples arrive, using a batch size 1 and learning rate $1\times10^{-7}$. This setting is conceptually introduced in~\cite{wang2021tent,wang2022cotta}, where the goal is to adapt to the local streaming data distribution. Offline TTT, by contrast, stores interaction samples in a buffer and performs batched optimization afterward. Within offline TTT, we further distinguish between per-task optimization and joint optimization over all tasks.
As shown in Figure~\ref{fig:ttt_analysis}, we find that fully online prompt updates are sensitive, leading to a large performance decrease.
Therefore, we use offline TTT as the primary setting and treat online TTT as an analysis of stability rather than the main deployment protocol.
Meanwhile, the difference between per-task and joint offline optimization is negligible.
These results suggest that batch size, rather than task grouping, is the key factor and may be important for maintaining meaningful latent representations during test-time training.
Stabilizing truly on-the-fly prompt updates remains an important future direction.

\bold{Learning-rate Sensitivity}
We also study learning-rate sensitivity for offline joint test-time training, since this basic hyperparameter has a large influence on performance.
A moderate learning rate of $10^{-5}$ performs best, outperforming both a larger update scale of $10^{-4}$ and a smaller update scale of $10^{-6}$.
This result suggests that prompt-only adaptation is not arbitrarily fragile, but it still benefits from using an update scale that is large enough to correct deployment-specific mismatch without destabilizing the learned prompt.

\begin{figure*}[t]
    \centering
    \includegraphics[width=\textwidth]{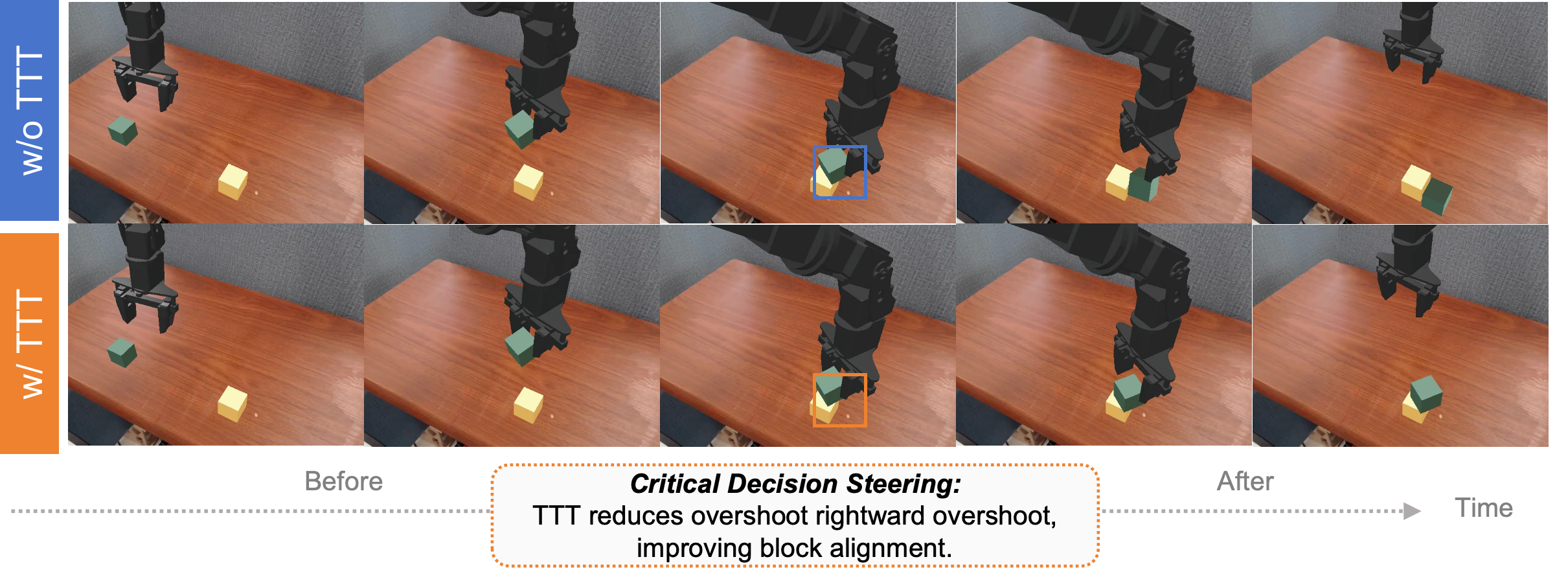}
    \caption{\textbf{Additional example of critical decision steering in the cube task.}
    This appendix figure complements Figure~\ref{fig:ttt_critical_decision} with another qualitative case.
    Before the critical placement decision, the trajectories with and without test-time training are similar.
    Near release, test-time training reduces rightward overshoot, improves block alignment, and leads to a successful stack.}
    \label{fig:ttt_critical_decision_cube}
\end{figure*}

\bold{Where the Gain Comes From.}
To obtain a valid comparison, we rerun experiments with the seed fully fixed\footnote{This is not feasible for standard evaluation because flow matching needs random sampling}.
This allows us to closely examine the cases in which test-time training improves behavior.
We find that most of these improvements follow a common pattern, which we term `critical decision steering'.
Figure~\ref{fig:ttt_critical_decision} illustrates one representative example.
Before the gripper grasps the carrot, the action outputs with and without TTT are nearly identical across all dimensions.
The divergence appears immediately after grasping: the model without TTT keeps moving downward, as if it fails to recognize that the carrot has already been secured and that the arm is blocked by the table, whereas the model with TTT instead lifts the arm appropriately.
From this decision point onward, the two rollouts separate and no longer overlap.
We observe similar behavior across tasks.
Many manipulation tasks consist of several stages, and transitions between stages often correspond to critical decision points.
One common example is deciding whether grasping has succeeded.
In the cube task, a different critical point appears during placement: after grasping the cube, the policy must decide whether the object is sufficiently aligned before release.
Appendix Figure~\ref{fig:ttt_critical_decision_cube} shows an additional example in which test-time training improves this placement decision by reducing overshoot and producing better block alignment.

\section{Conclusion}
We studied whether deployment-time improvement for Vision-Language-Action models can be achieved by updating a learned latent prompt.
To this end, we introduced TTT-VLA, a framework based on Latent Prompt Optimization that learns a latent prompt during training and performs prompt-only test-time training on a frozen policy at deployment time.
Experiments in SimplerEnv show that this design improves performance in both single-embodiment and multi-embodiment settings, including consistent gains from test-time training on WidowX and Google Robot tasks.

The empirical analysis further indicates that the benefit of test-time training often comes from correcting behavior at a small number of critical decision points, rather than globally changing the full action trajectory. A current limitation is that, under the present proxy task, these improvements mostly take the form of local corrections to erroneous actions near such critical points, rather than more complex corrective behaviors over longer horizons.
This may limit the ceiling of improvement that prompt-only test-time training can achieve. An important next step is to develop more effective prompt optimization procedures and to study how this interface scales with stronger VLA backbones. Additional limitations and open questions are discussed in Appendix~\ref{sec:systematic_limitations}.

\bibliographystyle{plainnat}
\bibliography{main}

\clearpage

\beginappendix

\begin{table*}[h]
    \centering
    \small
    \caption{\textbf{Extended comparison of deployment-time improvement strategies for VLA.}
    This appendix version expands the main-table comparison with additional practical considerations, including typical strengths and limitations of each strategy family. TTT-VLA belongs to the self-supervised TTT regime: it optimizes only a latent prompt while keeping the policy backbone frozen.}
    \label{app:teaser_comparison}
    \setlength{\tabcolsep}{6pt}
    \renewcommand{\arraystretch}{1.25}
    \resizebox{\textwidth}{!}{
        \begin{tabular}{p{0.15\textwidth}p{0.24\textwidth}p{0.24\textwidth}p{0.25\textwidth}}
            \toprule
            & \textbf{Human-Assisted Steering} & \textbf{RL Post-Training} & \textbf{Test-time Training } \\
            \midrule
            \textbf{Intervention interface} &
            Prompts, e.g., task-factorized instruction, metadata, trace &
            Typically policy itself &
            Typically parts of the parameters. In this paper, we only use latent prompts. \\
            \textbf{Improvement signal} &
            Human guidance / external signal&
            Reward &
            Self-supervised loss\\
            \textbf{Policy backbone status} &
            Frozen &
            Generally updated &
            Frozen \\
            \textbf{Typical strength} &
            Simple and controllable at deployment time &
            Can improve beyond the pretrained policy through continued interaction &
            Lightweight self-supervised improvement without updating the backbone \\
            \textbf{Typical limitation} &
            Depends on human or externally specified guidance and is often bounded by the pretrained policy &
            Usually costly, reward-dependent, and optimization-heavy &
            Depends on the quality of the proxy objective, and gains may remain local or bounded by the adaptation interface \\
            \textbf{Expected improvement pattern} &
            Mostly steers behavior within the capability envelope of the pretrained policy &
            Can reshape policy behavior more substantially through optimization &
            Often corrects deployment-specific mistakes while preserving most pretrained behavior \\
            \bottomrule
        \end{tabular}
    }
\end{table*}

\begin{table*}[h]
    \centering
    \small
    \caption{\textbf{Default test-time training configuration used in our experiments.}
    Unless otherwise stated, TTT-VLA uses the following setup for deployment-time prompt optimization.}
    \label{tab:ttt_config}
    \setlength{\tabcolsep}{10pt}
    \renewcommand{\arraystretch}{1.2}
    \begin{tabular}{lp{0.68\textwidth}}
        \toprule
        \textbf{Item} & \textbf{Value} \\
        \midrule
        TTT mode & Offline by default; online TTT is only used as an ablation \\
        Interaction buffer & 800 / $\sim$800 / $\sim$2000 for WidowX/GR VM/ GR VA \\
        Trajectory source & Rollouts collected from the pretrained policy \\
        Success / failure composition & Includes both successful and failed rollouts \\
        Optimization target & State grounding loss only \\
        Updated parameters & Latent prompt only \\
        Frozen parameters & Vision-language backbone, action expert, and state expert \\
        TTT steps & 500 for WidowX; 1000 for Google Robot \\
        Batch size & 128 \\
        Learning rate & $1\times10^{-5}$ \\
        Wall-clock cost & Approximately 15/30 minutes for WidowX/GR optimization \\
        Checkpoint selection & Fixed optimization horizon for each benchmark; no best-step selection on evaluation performance \\
        \bottomrule
    \end{tabular}
\end{table*}

\begin{table*}[t]
    \centering
    \footnotesize
    \caption{\textbf{Episode-aligned outcome comparison before and after TTT in the single-embodiment WidowX setting.}
    We align baseline and post-TTT rollouts by episode identity and compare binary task outcomes.
    The results show that TTT does not simply repair a fixed handful of cases; instead, it produces a net positive shift across paired rollouts, with both recovered failures and occasional regressions.}
    \label{tab:paired_episode_shift}
    \setlength{\tabcolsep}{5pt}
    \begin{tabular}{lrrrrrrrr}
        \toprule
        Task & Total & Base Succ. & Post Succ. & $\Delta$ & Succ.$\rightarrow$Fail. & Fail.$\rightarrow$Succ. & Always Succ. & Always Fail. \\
        \midrule
        Carrot & 200 & 139 & 149 & +10 & 28 & 38 & 111 & 23 \\
        Eggplant & 200 & 141 & 152 & +11 & 29 & 40 & 112 & 19 \\
        Spoon & 200 & 145 & 142 & -3 & 28 & 25 & 117 & 30 \\
        Cube & 200 & 83 & 96 & +13 & 29 & 42 & 54 & 75 \\
        \midrule
        Overall & 800 & 508 & 539 & +31 & 114 & 145 & 394 & 147 \\
        \bottomrule
    \end{tabular}
\end{table*}

\subsection{Systematic Limitations}\label{sec:systematic_limitations}

\paragraph{No real-world deployment evaluation.}
We do not yet evaluate TTT-VLA in fully real-world deployment.
This limitation should be understood in the context of the problem setting: unlike standard imitation-learning evaluation, real-world test-time training requires collecting fresh interaction data in the target environment before adaptation.
In this sense, the challenge is closer to real-world reinforcement-learning deployment, where interaction cost, system stability, and protocol design become part of the problem itself.
As test-time training for VLA is still at an early stage, many practical questions remain open, including how to structure data collection, how to control adaptation risk during deployment, and how to define stable real-world evaluation procedures.
We therefore view rigorous real-world TTT evaluation as an important but separate challenge beyond the present simulator-based study.

\paragraph{Limited proxy-task comparisons.}
In the current study, we instantiate Latent Prompt Optimization only with the state grounding proxy task.
Other proxy tasks may also be promising, including future-state prediction, future-image prediction, future representations prediction, or inverse-dynamics prediction.
Each of these alternatives may introduce different optimization challenges.
For example, future-state prediction may require care to avoid confusion with the main action-prediction objective. Future-image prediction may be challenges task and need a pretrained world model as a prediction head. Future representations prediction needs to find a effective encoder cross DINO, SigLIP, etc.
We focused first on establishing a stable and observable TTT setting for VLA, including basic hyperparameter choices, custom simulator evaluation rules, and personalized deployment-time optimization protocols.
This stabilization effort consumed substantial experimental time, so a broader proxy-task comparison remains future work.

\paragraph{No direct comparison to prior TTT methods.}
We do not provide a direct experimental comparison to prior TTT methods for manipulation policies.
For VLA specifically, the closest concurrent work is WorldAgen~\cite{worldagen_2026}, but its code is not publicly available and it does not report SimplerEnv results, making controlled comparison difficult.
Earlier work such as Self-Supervised Policy Adaptation during Deployment~\cite{pad_2020} studies deployment-time self-supervised adaptation, but it predates modern VLA models and is therefore not directly comparable in model class or evaluation setup.

\paragraph{Stability of online TTT.}
We observe that fully online prompt updates can be unstable when the optimization batch is small, sometimes leading to prompt collapse.
For this reason, we use offline TTT as the main deployment protocol in this paper.
Stabilizing online updates with proximal constraints, trust-region objectives, prompt-norm regularization, or micro-batch accumulation is an important direction for future work.

\paragraph{Deployment cost.}
TTT-VLA requires collecting interaction trajectories in the target environment and performing prompt optimization before evaluation.
Although only a small number of parameters are updated and the policy backbone remains frozen, the interaction and optimization cost is not zero.
We therefore report the number of trajectories, optimization steps, and wall-clock overhead explicitly, so that this trade-off is visible in addition to the final performance gain.

\begin{figure*}[h]
    \centering
    \includegraphics[width=\textwidth]{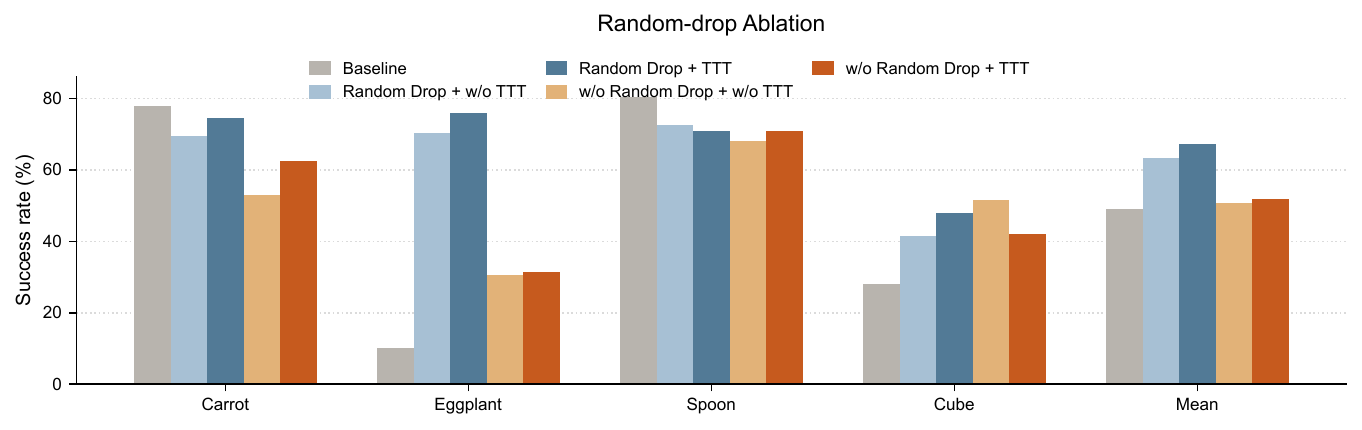}
    \caption{Ablation of random-drop training for latent prompt optimization in the single-embodiment setting. We compare the original baseline, the random-drop variants, and the corresponding variants without random drop across Carrot, Eggplant, Spoon, Cube, and the overall mean. Random-drop training consistently improves the corresponding variants, especially after test-time training.}
    \label{fig:random_drop_ablation}
\end{figure*}

\end{document}